# A fuzzy relation-based extension of Reggia's relational model for diagnosis handling uncertain and incomplete information

**Didier Dubois – Henri Prade**
Institut de Recherche en Informatique de Toulouse – C.N.R.S.
Université Paul Sabatier, 118 route de Narbonne
31062 TOULOUSE Cedex – FRANCE

## Abstract

Relational models for diagnosis are based on a direct description of the association between disorders and manifestations. This type of model has been specially used and developed by Reggia and his co-workers in the late eighties as a basic starting point for approaching diagnosis problems. The paper proposes a new relational model which includes Reggia's model as a particular case and which allows for a more expressive representation of the observations and of the manifestations associated with disorders. The model distinguishes, i) between manifestations which are certainly absent and those which are not (yet) observed, and ii) between manifestations which cannot be caused by a given disorder and manifestations for which we do not know if they can or cannot be caused by this disorder. This new model, which can handle uncertainty in a non-probabilistic way, is based on possibility theory and so-called twofold fuzzy sets, previously introduced by the authors.

## 1 INTRODUCTION

The paper views the diagnosis problem as it is considered in relation-based models where a relation describes the association between disorders and manifestations. This view, although elementary, enables us to discuss basic issues in relation with uncertainty in diagnosis problems. The completely informed case where there is no uncertainty in the association between disorders and manifestations and where all manifestations are observable and observed, is first dealt with in Section 2. Then a model is proposed in Section 3 for the case where we only have incomplete information about the manifestations which are present and about the manifestations which are indeed caused by a given disorder. This situation can be interpreted in terms of two-valued possibility and necessity measures. Namely we distinguish between manifestations whose presence is necessarily true (or if we prefer, certain) and those whose presence is only possible. The proposed model is compared in Section 4 to the parsimonious covering theory developed by Reggia et al. (1985) which appears to be a particular case. Section 6 presents a new model based on twofold fuzzy relations and twofold fuzzy sets (Dubois and Prade, 1987), which has a greater expressive power. As previously pointed out in Section 5, the model departs from fuzzy relational models first proposed by Sanchez (1977, 1979) and others (e.g. Tsukamoto and Terano, 1977; Pappis and Sugeno, 1985; Adlassnig et al., 1986; Asse et al., 1987; Kitowski and Bargiel, 1987) which are more appropriate when the intensity of the disorders and of the manifestations are a matter of degree. By contrast in our model, the presence of disorders or manifestations is not a matter of intensity but may be pervaded with uncertainty: they are either present or absent, but we may be more or less unsure about the presence of a manifestation when a disorder is present or about the observation of a manifestation. The new model presented in Section 6 is a graded version of the one proposed in the incompletely informed case but the handling of uncertainty remains ordinal and thus qualitative. Especially manifestations more or less certainly absent as well as those more or less certainly present are taken into account. Similarly the model manipulates the fuzzy set of manifestations which are more or less certainly produced by a disorder and the fuzzy set of manifestations which cannot be, more or less certainly, produced by this disorder. A preliminary version of this work appears in (Dubois and Prade, 1993)

## 2 RELATIONAL APPROACH : THE COMPLETELY INFORMED CASE

Let $\mathcal{S}$ be a system whose current state is described by means of a n-tuple of *binary* attributes $(a_1, ..., a_i, ..., a_n)$. When $a_i = 1$ we shall say that the manifestation $m_i$ is present; when $a_i = 0$, it means that $m_i$ is absent. When there is no manifestation present, $\mathcal{S}$ is said to be in its normal state and its state is described by the n-tuple $(0, ..., 0, ..., 0)$. Let $\mathcal{M}$ denote the set of the n possible manifestations $\{m_1, ..., m_i, ..., m_n\}$. Let $\mathcal{D}$ be a set of possible disorders $\{d_1, ..., d_j, ..., d_k\}$. A disorder can be present or absent. To each $d_j$ we associate the set $M(d_j)$ of manifestations which are entailed, or if we prefer caused, produced, by the presence of $d_j$ *alone*. We first consider the completely informed case where all the present manifestations are observed and where the set of manifestations which appear when a disorder is present is perfectly known. Thus if $m_i \notin M(d_j)$ it means that $m_i$ is not caused by $d_j$. We thus define a relation R on $\mathcal{D} \times \mathcal{M}$, defined by $(d_j, m_i) \in R \Leftrightarrow m_i \in M(d_j)$, which associates manifestations and disorders.

Given a set $M^+$ of present manifestations which are observed, the problem is to find what disorder(s) *may* have produced the manifestations in $M^+$. We suppose that the set $M^- = \mathcal{M} - M^+ = \overline{M^+}$ is the set of manifestations which are absent, i.e. all manifestations which are present are observed. While deductive reasoning enables us to predict the presence of manifestation(s) from the presence of disorder(s), abductive reasoning looks for possible cause(s) of observed effects. In other words, we look for plausible explanations (in terms of disorders) of an observed situation. Clearly while it is at least theoretically possible to find out all the possible causes which may have led to a given state of the system $\mathcal{S}$, the ordering of the possible solutions according to some levels of plausibility is out of the scope of logical reasoning, strictly speaking. However we may for instance prefer the solutions which involve a small number of disorders, and especially the ones, if any, which rely on only one disorder. This is called the principle of parsimony. In case several disorders may be jointly present (here we do not consider situations where disorder $d_i$ followed by $d_j$ has not the same effects in terms of manifestations as $d_j$ followed by $d_i$), we have to define the set of manifestations produced by the presence of a pair of disorders $(d_i, d_j)$ alone, and more generally by a tuple of disorders. In the hypothesis that effects can be added and do not interfere, we have

$$M(\{d_i, d_j\}) = M(d_i) \cup M(d_j) \quad (1)$$

and consequently

$$\overline{M(\{d_i, d_j\})} = \overline{M(d_i)} \cap \overline{M(d_j)}$$

i.e. the manifestations which are absent are those which are not produced by $d_i$ or $d_j$ separately. If this hypothesis is not acceptable, a subset $M(D)$ of entailed manifestations should be prescribed for each subset $D \subseteq \mathcal{D}$ of disorders which can be jointly present. In other words, we then work with a relation on $2^{\mathcal{D}} \times \mathcal{M}$, rather than on $\mathcal{D} \times \mathcal{M}$. If some disorders can never be jointly present, $2^{\mathcal{D}}$ should be replaced by the appropriate set $\mathcal{A}$ of associations of disorders which indeed make sense.

In the completely informed case described above, we have
i) $M^+ = \overline{M^-}$, i.e. all the present manifestations are observed, and equivalently all the manifestations which are not observed are indeed absent, and ii) $\forall d, M(d) = M(d)^+ = \overline{M(d)^-}$, where $M(d)^+$ (resp. $M(d)^-$) is the set of manifestations which are certainly present (resp. certainly absent) when disorder d alone is present. In case $M(D) \neq \bigcup_{d \in D} M(d)$, the condition (ii) above is supposed to hold $\forall D \in 2^{\mathcal{D}}$ in the completely informed case (and not only for $D = \{d\}$). Then the potential set $\widehat{D}$ of all the disorders which can individually be responsible for $M^+$ is given by

$$\widehat{D} = \{d \in \mathcal{D}, M(d) = M^+\}. \quad (2)$$

Note that $M(d) = M^+ \Leftrightarrow \overline{M(d)} = M^-$. Clearly, especially if $\widehat{D} = \emptyset$, we may be interested in the set $\widehat{DD}$ of subsets of disorders such that each subset may have caused $M^+$,

$$\widehat{DD} = \{D \in \mathcal{A} \subseteq 2^{\mathcal{D}}, M(D) = M^+\}. \quad (3)$$

Using the principle of parsimony, one might consider that the smaller the cardinality of D the more plausible it is. If $M(D)$ can be obtained as $\bigcup_{d \in D} M(d)$, then the set $D_0$ of disorders which alone *partially* explain $M^+$

$$D_0 = \{d \in \mathcal{D}, M(d) \subseteq M^+\} \quad (4)$$

may be of interest for building elements of $\widehat{DD}$. Clearly $D_0 \supseteq \widehat{D}$.

## 3 THE CASE OF INCOMPLETE INFORMATION

When not all the information is available, the set $M^+$ of manifestations which are certainly present and the set $M^-$ of manifestations which are certainly absent no longer form a partition of $\mathcal{M}$; indeed we have $M^+ \cap M^- = \emptyset$ but $M^+ \cup M^- \neq \mathcal{M}$. Similarly, for some d, we sometimes do not know if a manifestation m follows or not from d; in that case $m \notin M(d)^+$ and $m \notin M(d)^-$. In other words, the union of the set $M(d)^+$ of manifestations which are certainly produced by d alone and the set $M(d)^-$ of manifestations which certainly cannot be caused by d alone, no longer covers $\mathcal{M}$, i.e. $\exists d, M(d)^+ \cup M(d)^- \neq \mathcal{M}$, but, we always have $M(d)^+ \cap M(d)^- = \emptyset$. Denoting $M^0(d) = \mathcal{M} - (M^+(d) \cup M^-(d))$, if $m \in M^0(d)$ it means that m is only a possible manifestation of d. In particular m may be absent or present when d is present. Then d belongs to the set $\widehat{D}$ of potential disorders which alone can explain both $M^+$ and $M^-$ if and only if d does not produce with certainty any manifestation which is certainly absent in the evidence, and no observed manifestations must be ruled out by d. Formally we have

$$\widehat{D} = \{d \in \mathcal{D}, M(d)^+ \subseteq \overline{M^-} \text{ and } M(d)^- \subseteq \overline{M^+}\}. \quad (5)$$

This also writes

$\widehat{D} = \{d \in \mathcal{D}, M(d)^+ \cap \overline{M^-} = \emptyset \text{ and } M(d)^- \cap \overline{M^+} = \emptyset\}$. (5A)

Clearly (5) reduces to (2) in the completely informed case since then $\overline{M^-} = M^+$ and $M(d)^- = \overline{M(d)^+}$. Note that if $M(d)^+ = \emptyset = M(d)^-$, then $d \in \widehat{D}$, i.e. a disorder for which absolutely no knowledge is available about its effects, can be always considered as a potential responsible for observed manifestations. Note also that (5) satisfactorily covers the extreme situation where there is no genuine disorder. Indeed if $d \in \widehat{D}$, $M(d)^- = \mathcal{M}$ entails $M^+ = \emptyset$, i.e. a "disorder" without manifestation cannot explain a situation where a manifestation is observed; reciprocally $M^- = \mathcal{M}$ entails $M(d)^+ = \emptyset$, if $d \in \widehat{D}$, i.e. if we are certain that there is no manifestation, this is only compatible with a "disorder" which is not certainly followed by a manifestation.

When $\widehat{D} = \emptyset$, we can look for explanations in terms of subsets of disorders which are not singletons. (3) is then extended by

$\widehat{DD} = \{D \in \mathcal{A} \subseteq 2^{\mathcal{D}}, M(D)^+ \subseteq \overline{M^-} \text{ and } M(D)^- \subseteq \overline{M^+}\}$ (6)

for the subsets of disorders which alone may explain $M^+$ and $M^-$. As expected what is present and what is absent play symmetrical roles, exchanging + and − in (5) or (6). Note that if $M^- = \emptyset$, i.e. if we only know manifestations which are certainly present, (5) (or (6)) may yield a result $\widehat{D} \neq \mathcal{D}$ (or $\widehat{DD} \neq \mathcal{A}$) provided that $\overline{M(d)^-} \neq \mathcal{M}$, i.e. we have non-trivial information on the set of manifestations $M(d)^-$ (or $M(D)^-$) which may be produced by a disorder $d$ (or a subset of disorders $D$) alone; indeed $\overline{M(d)^-} \supseteq M(d)^+$ (resp. $\overline{M(D)^-} \supseteq M(D)^+$) gathers the manifestations which are certainly produced by $d$ (resp. $D$) and the manifestations for which we do not know if they can or cannot follow from $d$ (resp. $D$), i.e. if $M^- = \emptyset$,

$\widehat{D} = \{d \in \mathcal{D}, M(d)^- \subseteq \overline{M^+}\}$.
$\widehat{DD} = \{D \in \mathcal{A} \subseteq 2^{\mathcal{D}}, M(D)^- \subseteq \overline{M^+}\}$ (7)

In the non-completely informed case the hypothesis (1) that effects can be added and do not interfere writes (for two disorders)

$$\begin{cases} M(\{d_i, d_j\})^+ = M(d_i)^+ \cup M(d_j)^+ \\ \text{and} \\ M(\{d_i, d_j\})^- = M(d_i)^- \cap M(d_j)^-. \end{cases}$$ (8)

Clearly (8) reduces to (1) in the completely informed case. Note that the second equality of (8) still writes

$\overline{M(\{d_i, d_j\})^-} = \overline{M(d_i)^-} \cup \overline{M(d_j)^-}$

which says that the possible manifestations of two simultaneous disorders gather the manifestations possibly produced by each disorder, as for certain manifestations.

## 4 LINK WITH REGGIA ET AL.'S APPROACH

Reggia et al. (1985) (see also Peng and Reggia, 1990) have extensively studied a relation-based formulation of diagnosis problems. In their model they assume the knowledge of a relation between disorders and manifestations, such that the fact that the pair $(d_j, m_i)$ satisfies this relation "means $d_j$ may directly cause $m_i$. Note that this does *not* mean that $d_j$ *necessarily* causes $m_i$, but only that it *might*", as stated in (Peng and Reggia, 1990). But these authors do not explain why they make this choice for interpreting the "causal" relation between $\mathcal{M}$ and $\mathcal{D}$. Moreover the set $M^+$ of "manifestation known to be present" is supposed to be available.

Thus, in their model, what is known is the set $\overline{M(d)^-}$ of manifestations possibly attached to a disorder, for each disorder $d$; it is also assumed that (8) holds for computing $\overline{M(D)^-}$ for $D \subseteq \mathcal{D}$. Since in this model $M^- = \emptyset$, (7) applies, and indeed $\{d\}$ such that $\overline{M(d)^-} \supseteq M^+$ and more generally $D$ such that $\overline{M(D)^-} \supseteq M^+$ (applying (8), i.e. $\overline{M(D)^-} = \bigcup_{d \in D} \overline{M(d)^-}$) are called "*covers*" of $M^+$ by Peng and Reggia (1990).

These authors more particularly look for so-called "*parsimonious*" covers, especially relevant, irredundant and minimum covers. $D$ is a *relevant* cover if $\forall d \in D, \exists m \in M^+, (d, m) \in R$; $D$ is *irredundant* if none of its proper subsets is also a cover of $M^+$; $D$ is *minimum* if its cardinality is smallest among all covers of $M^+$. Clearly a one-disorder cover is a minimal cover, a minimal cover is an irredundant cover, an irredundant cover is a relevant cover, a relevant cover is obviously a cover. The set of relevant covers is defined by

$DD^* = \{D \in \mathcal{A} \subseteq 2^{\mathcal{D}}, \forall d \in D, \overline{M(d)^-} \cap M^+ \neq \emptyset\}$ (9)

since in Peng and Reggia (1990) the relation $R$ is defined by $(d, m) \in R \Leftrightarrow m \in \overline{M(d)^-}$, using our notations. These notions could easily be extended to framework where possible and sure manifestations are told apart. For instance the set of relevant covers could be defined by $\{D \in \mathcal{A} \subseteq 2^{\mathcal{D}}, \overline{M(d)^-} \cap M^+ \neq \emptyset \text{ and } \overline{M(d)^+} \cap M^- \neq \emptyset\}$, weakening the conditions $M(D)^+ \subseteq \overline{M^-}$ and $M(D)^- \subseteq \overline{M^+}$ in (6). Consequently our relational diagnosis model is more general than Reggia's, in the incompletely informed case.

## 5 GRADED UNCERTAINTY VS. GRADED INTENSITY OF PRESENCE

In the late seventies Sanchez (1977, 1979), Tsukamoto and Terano (1977) already developed diagnosis methods based on a fuzzy relational model. Several slightly different proposals have been made. We consider here the simplest version which corresponds to the type of diagnosis problem presented in the preceding section, where, i) R is a fuzzy relation defined on $\mathcal{D} \times \mathcal{M}$, and ii) $M^+$ is a fuzzy set. Indeed, it seems desirable in practice to be able to model a more gradual association between manifestations and disorders, and to take into account the uncertainty or vagueness pervading the observation of manifestations. Although this suggests the possible interest of some fuzzy set-based methods, it is not right away clear what should be the precise interpretation of the degrees of association $\mu_R(d_j, m_i)$, of the degrees of membership $\mu_{M^+}(m_i)$ when the relation R or the subset $M^+$ is fuzzy, or of the degrees attached to disorders which are then proposed for the explanation of $M^+$, knowing R. For instance Kitowski and Bargel (1987) spoke of the partial occurrence of a disorder, or of the uncertain observation of a manifestation, but other interpretations can be thought of.

There are basically two types of interpretations which can be contemplated, one pertaining to the level of fulfilment of a gradual manifestation, the other to the uncertainty of statements pertaining to observations. Namely, on the one hand we may be uncertain about the presence of a manifestation because its observation is, difficult, costly, or even untimely, or on the other hand manifestations may be a matter of degree (e.g. in medical diagnosis, the fever of a patient may be more or less high). In the first interpretation the uncertainty pervading our knowledge of actual manifestations stems from the use of easy, or quick observation methods whose results are imprecise. This may occur even if the presence or absence of a manifestation is an all or nothing matter: the presence of a binary manifestation can be more or less certain ($\mu_{M^+}(m_i) > 0$) or more or less possible ($1 - \mu_{M^-}(m_i) > 0$) due to unreliable or imprecise measurement. This case is studied in the following.

On the contrary, $\mu_{M^+}(m_i)$ can model the *intensity* of a manifestation $m_i$. This second interpretation is only possible if the underlying attribute $a_i$ is no longer binary. For instance, instead of considering that $m_i$ means "fever" or "not fever", it looks more natural to consider "high fever" as a manifestation. The vagueness pertaining to the word "high" reflects the fact that fever is a matter of degree and can be measured on a numerical scale S (ranging from 35 to 42 degrees, say). "High fever" refers to a fuzzy set F of S (that means "close to 40", say) in a given context. If $x \in S$ is the temperature of the patient, then $\mu_{M^+}$(high fever) = $\mu_F(x) = 1 - \mu_{M^-}$(high fever), since $M^-$ refers to the absent manifestations and $\mu_{M^-}$(high fever) = 1 means "not high fever". The semantics of the fuzzy complementation "1 - " significantly differs in the two situations: when grades of membership express intensities of presence of symptoms the complementation is dictated by an equivalence between strong intensity of "high fever" and low intensity of "not-high fever". When grades of membership express degrees of uncertainty of presence of binary symptoms, the complementation reflects the duality between certainty and possibility of presence, and obeys the following rule: what is cer<u>tain</u> must be possible, but not the converse, hence $M^+ \subseteq M^-$.

Similarly the degree of association $\mu_R(d_j, m_i)$ may account for the uncertainty that $m_i$ follows from the presence of $d_j$, or for the intensity of manifestation $m_i$ when $d_j$ is present (which again supposes that the severeness of the effect can be graded); if the level of disorder $d_j$ may itself be more or less severe $\mu_R(d_j, m_i)$ may even correspond to a gradual rule, like the more severe the disorder, the stronger the manifestation. The intensity of the disorder then simply reflects the level of matching between (numerical) measurements and the most acute forms of the manifestations caused by the disorder (like "high fever").

On the contrary if we interpret $\mu_R(d_j, m_i)$ as a degree of uncertainty it may either mean to what extent $m_i$ *necessarily* follows from the presence of $d_j$ or only to what extent it is *possible* that $m_i$ is present when $d_j$ is present. In particular if the fuzzy set $M(d_j)$ ($\mu_{M(d_j)}(m_i) = \mu_R(d_j, m_i)$) gathers the manifestations which more or less certainly or necessarily follow from $d_j$, the fact that $m_i$ is such that $\mu_R(d_j, m_i) = 0$ means that either $m_i$ does not follow from $d_j$, or that we do not know (i.e. we have absolutely no certainty that $m_i$ is caused by $d_j$); by contrast if $M(d_j)$ is the fuzzy set of manifestations which more or less possibly follows from $d_j$, $\mu_R(d_j, m_i) = 0$ means that $m_i$ cannot be caused by $d_j$, but $\mu_R(d_j, m_i) = 1$ expresses no certainty that $m_i$ *should* accompany $d_j$, it only expresses that it is fully possible. If $\mu_R(d_j, m_i)$ does reflect the uncertainty of presence of manifestation $m_i$ due to disorder $d_j$, but the end-points of the unit interval are interpreted as impossible (0) and certain (1), then it is difficult to imagine that $\mu_R(d_j, m_i)$ represents something else than the probability that $d_j$ causes $m_i$ when $d_j$ is present, as used by Peng and Reggia (1990), this probability being subjective and objective. According to the interpretation we have in mind, it will lead to different models with different interpretations of the results. To some extent the literature on fuzzy relational equations for diagnosis suffers from a lack of concern for these representational issues.

The reader is referred to Dubois and Prade (1992b) for a detailed study and discussion of proposals made in the fuzzy set literature for handling the diagnosis problem on the basis of fuzzy relation (in)equations. Viewed in retrospect these proposals are more appropriate for dealing with the case where the intensity of the disorders and of the manifestations is graded, even if they have been often mistakenly used for dealing with uncertainty. In the following we extend the model presented in Section 3 to the case where the uncertainty is graded. We use the possibilistic framework which offers an ordinal view of uncertainty only requiring the comparison of levels of uncertainty.

## 6 A NEW MODEL BASED ON TWOFOLD FUZZY SETS

In this section, we propose a graded counterpart of the model presented for the non-completely informed case. Namely $M^+$ and $M^-$ are now fuzzy sets of manifestations which are more or less certainly present, and more or less certainly absent respectively. However we keep the requirement $M^+ \cap M^- = \emptyset$ (where the intersection is defined by the min operation), i.e. we cannot be somewhat certain both of the presence and of the absence of the same manifestation simultaneously. Similarly, $M(d)^+$ (more generally $M(D)^+$) and $M(d)^-$ (more generally $M(D)^-$) will denote the fuzzy sets of manifestations which are respectively more or less certainly present and more or less certainly absent when disorder d alone is present (more generally when the subset D of disorders is present). Obviously, we also assume $\forall d, M(d)^+ \cap M(d)^- = \emptyset$ (and $\forall D, M(D)^+ \cap M(D)^- = \emptyset$).

By complementation (defined by $\mu_{\overline{F}} = 1 - \mu_F$), we obtain the fuzzy sets $\overline{M^-}$, $\overline{M(d)^-}$ (and more generally $\overline{M(D)^-}$) of manifestations which are more or less possibly present, respectively, in the considered situation, when d is present, or when disorders in D are altogether present. This corresponds to the usual duality between what is (more or less) certain, i.e. necessarily true, and what is (more or less) possibly true. Indeed a pair of dual possibility and necessity measures $\Pi$ and $N$ are related by the relation $\Pi(A) = 1 - N(\overline{A})$, for any event A (here A represents the presence of a manifestation); see Dubois and Prade (1988a) for instance.

Note that $M^+ \subseteq \overline{M^-}$, $M(d)^+ \subseteq \overline{M(d)^-}$, and $M(D)^+ \subseteq \overline{M(D)^-}$ in the sense of fuzzy set inclusion. An even stronger inclusion holds. Since $M^+ \cap M^- = \emptyset$, we have

$$\{m_i \in \mathcal{M}, \mu_{M^+}(m_i) > 0\} \subseteq \{m_i \in \mathcal{M}, \mu_{M^-}(m_i) = 0\}$$
$$= \{m_i \in \mathcal{M}, \mu_{\overline{M^-}}(m_i) = 1\} \quad (10)$$

i.e. the support of $M^+$ is included in the core of $\overline{M^-}$; the same holds for $M(d)^+, \overline{M(d)^-}$, or $M(D)^+, \overline{M(D)^-}$. This is in agreement with the fact that for crisp events A, we have $N(A) > 0 \Leftrightarrow \Pi(\overline{A}) < 1 \Rightarrow \Pi(A) = 1$ since then one of A or $\overline{A}$, at least, should be completely possible; see Dubois and Prade (1988a) for instance.

A pair of fuzzy sets (F,G) such that $F \cap \overline{G} = \emptyset$ is called a twofold fuzzy set (Dubois and Prade, 1987). Twofold fuzzy sets (F,G) have been introduced for modelling incompletely known sets, i.e. sets for which we know elements gathered in F, which more or less certainly belong to it, as well as other elements, gathered in $\overline{G}$, which more or less certainly do not belong to it. But $F \cup \overline{G}$ may not cover the whole referential. Similarly, the pairs $(M(d)^+, \overline{M(d)^-})$ and $(M(D)^+, \overline{M(D)^-})$ define twofold fuzzy relations on $\mathcal{D} \times \mathcal{M}$, and $2^{\mathcal{D}} \times \mathcal{M}$ respectively.

The extension to fuzzy sets of equation (5) and (6) can be very simply done on the basis of (5A). It requires that the extent to which two fuzzy sets F and G of $\mathcal{M}$ intersect be evaluated. The consistency between F and G is simply defined as (Zadeh, 1979)

$$\text{cons}(F,G) = \sup_{m \in \mathcal{M}} \min(\mu_F(m), \mu_G(m)).$$

It computes the degree of existence of some common element for F and G. (5A) is based on checking the inconsistency level between fuzzy sets, that is $1 - \text{cons}(F,G)$. The fuzzy extension of (5) then leads to compute a fuzzy set $\widehat{D}$ of plausible unique disorders as $\forall d \in \mathcal{D}$:

$$\mu_{\widehat{D}}(d) = \min(1 - \text{cons}(M(d)^+, M^-), 1 - \text{cons}(M(d)^-, M^+))$$
$$= 1 - \max(\text{cons}(M(d)^+, M^-), \text{cons}(M(d)^-, M^+)) \quad (11)$$

where the minimum operator expresses the conjunction of the conditions in (5A).

*Remark* : Because in the model, we assume $M^+(d) \cap M^-(d) = \emptyset$, the strong inclusion (10) follows between $M^+(d)$ and $\overline{M^-(d)}$. If we evaluate (5) using the strong inclusion, we must use the following inclusion index of F in G :

$$\text{inc}(F,G) = \inf_{m \in \mathcal{M}} \max(1 - \mu_F(m), \mu_G(m)). \quad (12)$$

Indeed, $\text{inc}(F,G) = 1 \Leftrightarrow \text{support}(F) \subseteq \text{core}(G) \Leftrightarrow F \cap \overline{G} = \emptyset$. The implication $\max(1 - a, b)$ is well-known in multiple valued logic as Dienes implication, and multiple valued implications are basic for building fuzzy set inclusion indices (Bandler and Kohout, 1980). It is easy to check that (11) can be written using (12) as

$$\mu_{\widehat{D}}(d) = \min(\text{Inc}(M(d)^+, \overline{M^-}), \text{inc}(M(d)^-, \overline{M^+}))$$

since inc(F,G) = 1 − cons(F,Ḡ). □

(11) clearly expresses that a disorder d is all the less a candidate explanation as the fuzzy set of its more or less certain effects overlaps the fuzzy set of manifestations more or less certainly absent, or as the fuzzy set of effects which are more or less certainly absent when d is present overlaps the fuzzy set of manifestations which are more or less certainly present. This is intuitively satisfying.

In the particular case where $M^- = \emptyset$, i.e. we only have positive observations, $\widehat{D}$ is defined $\forall d \in \mathcal{D}$ by

$$\mu_{\widehat{D}}(d) = \min_{i=1,n} \max(1-\mu_{M^+}(m_i), \mu_{\overline{M(d)^-}}(m_i)). \quad (13)$$

Note that the modelling of uncertainty remains here qualitative. Indeed, we could use a finite completely ordered chain of levels of certainty ranging between 0 and 1, i.e. $\ell_1 = 0 < \ell_2 < ... < \ell_n = 1$ instead of [0,1], with $\min(\ell_i, \ell_k) = \ell_i$ and $\max(\ell_i, \ell_k) = \ell_k$ if $i \leq k$, and $1 - \ell_i = \ell_{n+1-i}$.

Taking into account the incomplete nature of the information about the presence or absence of manifestations decreases the discrimination power when going from the completely informed case (equation (2)) to the incomplete information case (equations (5) or (5A)), since then the number of possible disorders in $\widehat{D}$ increases. This is due to the fact that now there are manifestations which are neither certain nor impossible and consequences of the presence of a given disorder d which are only possible, as pictured in Figure 2 while $M(d)^+ = M(d)$ and $M(d)^- = \overline{M(d)}$ on Figure 1 (similar figures can be drawn for $M^+$ and $M^-$).

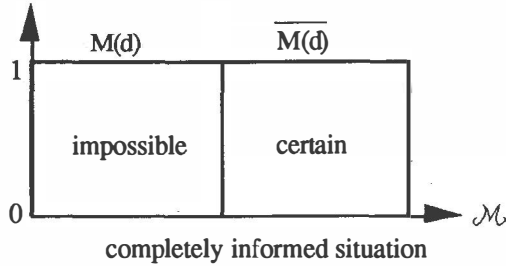

completely informed situation

Figure 1

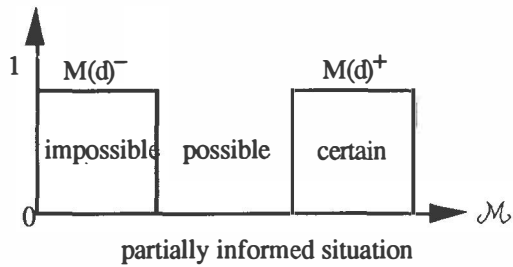

partially informed situation

Figure 2

This suggests that, in order to improve the discrimination power of the model, we have to refine the non-fuzzy model in such a way that consequences (resp. manifestations) previously expressed as certain (resp. certainly present) and impossible (resp. certainly absent) remain classified in the same way and where some possible consequences (resp. possibly present manifestations) are now allowed to be either somewhat certain (resp. somewhat certainly present) or somewhat impossible (resp. somewhat certainly absent). See Figure 3. Then (12) enables us to rank-order the possible disorders which are compatible with the observations. This counterbalances the increase of candidates due to the incompleteness of the information.

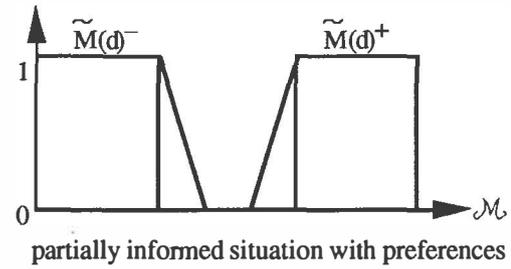

partially informed situation with preferences

Figure 3

It can be easily shown that adding preference levels on top of incompleteness modeling can at the same time enable the discrimination power of the completely informed situation to be recovered, and also enable the extra disorders obtained on the partially informed situation to be ranked in terms of their plausibility levels. Indeed, assume that the dichotomy $M(d), \overline{M(d)}$ of Figure 1 represents a first approximation of the fuzzy sets $\widetilde{M}(d)^+$ and $\widetilde{M}(d)^-$ of Figure 3 in the sense that

$$M(d) = \text{support}(\widetilde{M}(d)^+) \;;\; \overline{M(d)} = \text{support}(\widetilde{M}(d)^-)$$

and moreover $M(d)^+ = \text{core}(\widetilde{M}(d)^+)$, $M(d)^- = \text{core}(\widetilde{M}(d)^-)$. Then it can be proved that both results of the completely informed and partially informed situations can be recovered by the fuzzy model. Namely let $\widehat{D}_c, \widehat{D}_p$, and $\widehat{D}$ be the sets of disorders obtained via equations (2), (5) and (11) respectively. We shall assume non-fuzzy incomplete observations, in which case (2) becomes $\widehat{D}_c = \{d \in \mathcal{D}, M^+ \subseteq M(d) \subseteq \overline{M^-}\}$. Then we can show the following equalities,

**Proposition** : $\widehat{D}_c = \text{core}(\widehat{D})$ ; $\widehat{D}_p = \text{support}(\widehat{D})$

**Proof** :
$d \in \text{core}(\widehat{D})$
$\Leftrightarrow \text{cons}(\widetilde{M}(d)^+, M^-) = \text{cons}(\widetilde{M}(d)^-, M^+) = 0$ (from (11))

$$\Leftrightarrow \widetilde{M}(d)^+ \cap M^- = \emptyset = \widetilde{M}(d)^- \cap M^+$$
$$\Leftrightarrow M(d) \cap M^- = \emptyset, \overline{M(d)} \cap M^+ = \emptyset$$
$$\Leftrightarrow M^+ \subseteq M(d) \subseteq \overline{M^-} \Leftrightarrow d \in \widehat{D}_c$$

$d \in \text{support}(\widehat{D})$
$$\Leftrightarrow \text{cons}(\widetilde{M}(d)^+, M^-) < 1 \text{ and } \text{cons}(\widetilde{M}(d)^-, M^+) < 1$$
$$\Leftrightarrow M(d)^+ \cap M^- = \emptyset \text{ and } M(d)^- \cap M^+ = \emptyset$$
$$\Leftrightarrow d \in \widehat{D}_p. \qquad \text{Q.E.D.}$$

This result indicates that our approach, although much more qualitative than one based on probability theory, still possesses the ability to rank-order the set of plausible unique disorders explaining an incomplete set of manifestations.

Clearly (11) and (13) straightforwardly extend to subsets D of disorders which altogether explain both $M^+$ and $M^-$, substituting D to the singleton {d} in (11) and (13). Indeed when $\widehat{D} = \emptyset$, we have to look for two-element subsets D which may account for $M^+$ and $M^-$, and then for three-element subsets if there is no two-element one, and so on until a plausible explanation is found. The decomposition properties (8) when they hold, easily extend to the fuzzy case under the form

$$\begin{cases} \mu_{M(\{d_i,d_j\})^+} = \max(\mu_{M(d_i)^+}, \mu_{M(d_j)^+}) \\ \mu_{M(\{d_i,d_j\})^-} = \min(\mu_{M(d_i)^-}, \mu_{M(d_j)^-}). \end{cases} \qquad (14)$$

Note that (14) is coherent with the definition of the union of twofold fuzzy sets. However, if we look for multiple disorders explaining a given set of manifestations, it is clear that we shall have a trade-off problem between small sets of disorders which are most plausible in the sense of the parsimony principle, and bigger sets of disorders which are more plausible because they ensure a better covering of the observed manifestations. This topic, along with the semantics of (14) for the representation of independent disorders, requests further investigation.

## 7 CONCLUDING REMARKS

In this paper we have proposed a new model for diagnosis problems, which is more expressive than Reggia's pure relational model for representing the available causal information. The application of this model to practical diagnosis problems is currently under investigation (Cayrac et al., 1993). Other models allowing for non-binary attributes for expressing the intensity of manifestations and for the expression of gradual association between the intensities of disorders and manifestations have still to be developed.

In spite of its greater expressiveness, the model we have proposed here has still several limitations. Let us mention two of them. The relational model we consider associates directly disorders and manifestations. More generally we may have two relations, between $\mathcal{D}$ and an intermediary set $\mathcal{S}$, and between $\mathcal{S}$ and $\mathcal{M}$; see (Peng and Reggia, 1987) on this point. Besides, we are not able to capture the most general kind of incomplete information. For instance we cannot express that we are certain that manifestations $m_i$ or $m_j$ are present (but perhaps one of them is absent), or a similar information stating that when disorder d is present it is certain that $m_i$ or $m_j$ are present (and that $m_k$ or $m_\ell$ are absent) for instance. See Dubois and Prade (1988b) for the modelling of such pieces of knowledge in case of graded uncertainty. The treatment of the most general kind of incomplete information would require to work with a (fuzzy) relation R defined on $2^{\mathcal{D}} \times 2^{\mathcal{M}}$.

In the above model all the effects of a disorder are assumed to take place simultaneously. This is not always the case in practice (e.g. Console and Torasso, 1991) and it may be more realistic to associate with a disorder the sets of manifestations $M(d)_t^+$ and $M(d)_t^-$ which are respectively more or less certainly present and more or less certainly absent t time units after that the disorder begins to take place. More discussion along this line is in (Dubois and Prade, 1993)

Another topic for further research is the expression in a logical formalism of the proposed approach (as Reiter (1987)'s logical model encodes Reggia's basic ideas), in order to relate it with methods based on possibilistic assumption-based truth-maintenance systems; see (Benferhat, Dubois, Lang and Prade, 1992; Dubois and Prade, 1992a). More generally it would be interesting to develop a logical framework where it would be possible to express both weighted deductive rules associating manifestations to disorders and weighted evocation rules (in the sense of Pearl (1988)) associating possible disorders to manifestations, and perform local reasoning tasks.

Lastly an interesting issue to be investigated later on is how to take a priori information about disorders into account in the framework of non-probabilistic relational models of diagnosis. To-date the only available framework for modeling a priori information in diagnosis problems is Bayesian probability. However, it is well-known that a full-fledged probabilistic prior is not always available. This does not mean that no a priori knowledge is present. The framework of possibilistic abduction (Dubois and Prade, 1992a) may offer a tool for modeling non-probabilistic priors either in a purely ordinal setting or in a quantitative but less demanding framework than Bayesian probability. The ultimate aim of this research could be the design of a general causal model of diagnosis under uncertain and incomplete information that encompasses the Bayesian model as a special case (when prior probabilities are likelihood functions for the description of the causal knowledge) and the ordinal model described in this paper (when the prior knowledge corresponds to total ignorance). In such a general framework, it is clear that the entries of the fuzzy causal relations should be interpreted in terms of conditional uncertainty measures that should be more general than both probability and possibility measures (like belief functions or probability bounds).